\documentclass[10pt,twocolumn,letterpaper]{article}

\usepackage{cvpr}
\usepackage{times}
\usepackage{epsfig}
\usepackage{graphicx}
\usepackage{amsmath}
\usepackage{amssymb}
\usepackage{comment}
\usepackage{color}
\usepackage{subfigure}
\usepackage{multirow}

\usepackage[breaklinks=true,bookmarks=false]{hyperref}

\cvprfinalcopy 

\begin{document}
\title{ESA-ReID: Entropy-Based Semantic Feature Alignment for Person re-ID} 

\author{
Chaoping Tu\\
Alibaba Group\\
Beijing China\\
{\tt\small chaoping.tcp@alibaba-inc.com}
\and 
Yin Zhao\\
Alibaba Group\\
Beijing China\\
{\tt\small yinzhao.zy@alibaba-inc.com}
\and
Longjun Cai\\
Alibaba Group\\
Beijing China\\
{\tt\small longjun.clj@alibaba-inc.com}
}

\maketitle

\begin{abstract}
Person re-identification (re-ID) is a challenging task in real-world. 
Besides the typical application in surveillance system, re-ID also has significant values to improve the recall rate of people identification in content video (TV or Movies). However, the occlusion, shot angle variations and complicated background make it far away from application, especially in content video. 
In this paper we propose an entropy based semantic feature alignment model, which takes advantages of the detailed information of the human semantic feature.
Considering the uncertainty of semantic segmentation, we introduce a semantic alignment with an entropy-based mask which can reduce the negative effects of mask segmentation errors. We construct a new re-ID dataset based on content videos with many cases of occlusion and body part missing, which will be released in future. Extensive studies on both existing datasets and the new dataset demonstrate the superior performance of the proposed model.

\end{abstract}

\section{Introduction}
Person re-identification (re-ID), which aims to match people across different cameras with different viewpoints, is a very challenging task in real-world. Ubiquitous occlusion, complex background, illumination variations jointly bring the problem even harder. 
Most of the public datasets, such as Market-1501\cite{datamarket}, DukeMTMC\cite{dataduke1,dataduke2}, are based on surveillance video, which empowers the person re-ID methods to achieve great progress in recent years.
However, there is hardly any research working on person re-ID in TV or movie videos (which will be called content video in the following).
The content video is a different domain from surveillance video. 
It is an artifact and manipulated by humans (cameraman or director) while the surveillance video captures the natural world with inartificial time flow and constant view point. 
The artificial manipulations in content video will bring more people occlusions, body part missing, angle and scale variations.

\begin{figure*} [htbp]
\centering
\label{fig:content_example}
\subfigure[]{
\begin{minipage}[h]{0.45\linewidth}
\label{subfig:occlusion}
\centering 
\includegraphics[width=3.0in]{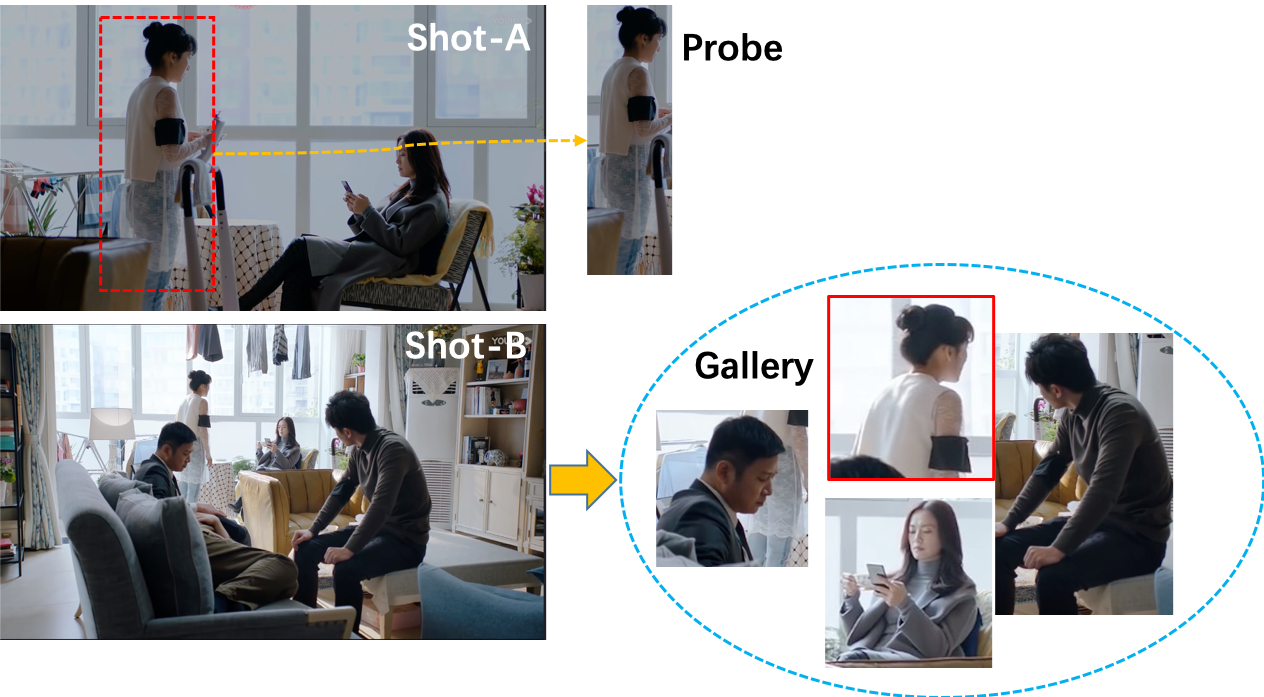} 
\end{minipage}
}
\hspace{10mm}
\subfigure[]{
\begin{minipage}[h]{0.45\linewidth} 
\label{subfig:recognition}
\centering 
\includegraphics[width=3.0in]{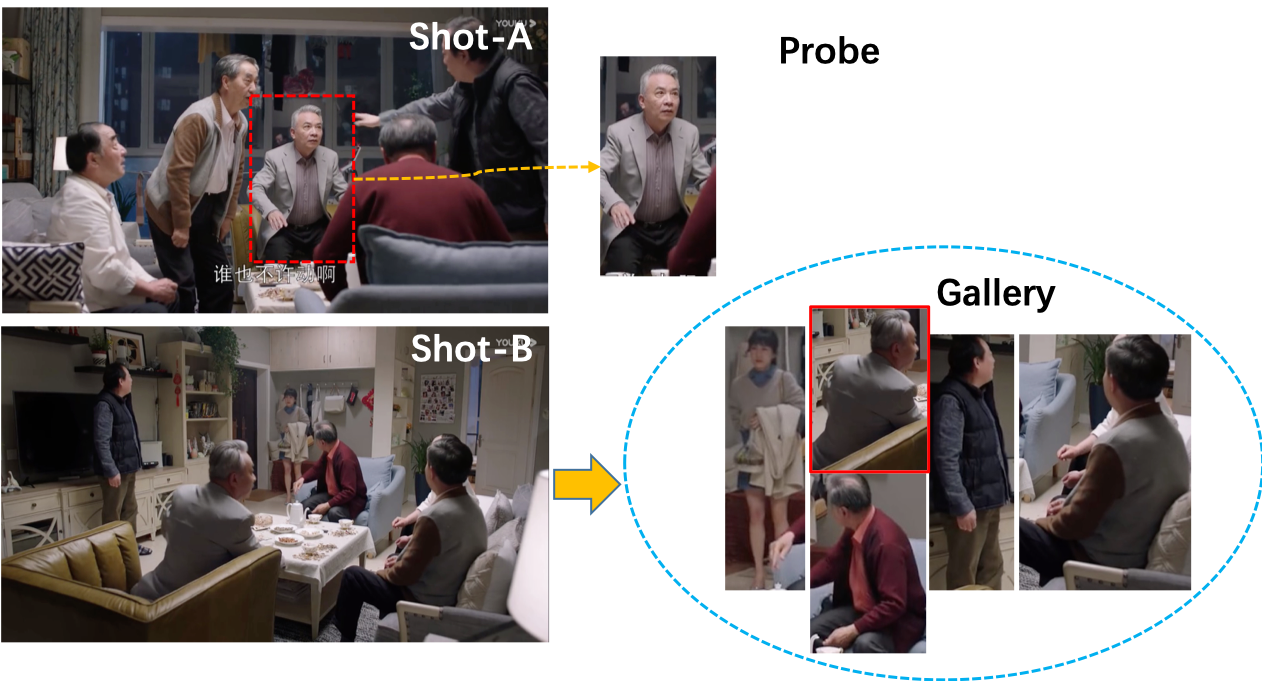} 
\end{minipage}
}
\caption{Typical examples of content video. People in the content videos have large variations in angles, scales and occlusion.}
\end{figure*}
An example of content video can be seen in Fig.\ref{fig:content_example}.
In Fig.\ref{subfig:occlusion}, A and B are two consecutive shots describing the same scene but with different shot angles and camera distances. Matching the people in shot A to shot B can be regarded as a re-ID problem for half body and whole body.
Re-identifying the person in the content video, especially matching the people in the shot with ill-conditioned for recognition or tracking to the other shot, which is ready to identify, is much more helping to improve the recall rate. Fig.\ref{subfig:recognition} shows an example, where the detected people in shot A has frontal face and can be recognized. If we could re-identify the same person in shot B, we could recognize the character even if the person is side or back to the audience.

In this paper we want to solve the re-ID problem in both surveillance and content videos, focusing on the challenges of occlusion and body part missing.
Those cases are much more remarkable in content videos. 
This problem is generally defined as partial re-ID or occluded re-ID, where the
person in the probe or gallery is occluded or partially captured. 
Several methods\cite{he2018DSR,he2018SFR,miao2019PGFA} have investigated this problem and designed many methods to align the two peoples by static or dynamic local features.
However, due to the lack of dedicated high-quality dataset for occlusion and body part missing, few methods have studied the alignment from semantic level in depth, not alone considering the uncertainty of semantic segmentations which would affect the alignment effects.

We propose a new re-ID model called entropy-based semantic alignment re-ID(ESA-ReID), dedicated for the occlusion, missing body part issues in both content video and surveillance video. 
This model highly utilizes the detailed human body sematic feature and its uncertainty. 
We perform a semantic alignment of human semantic features based on visible score and confidence score defined on entropy.
It will dynamically align the common and confidence semantic human regions without additional computation load. 
We also construct a new dataset called Drama-ReID, which, as far as we know, is the largest partial re-ID dataset retrieved from drama or movies. 
Extensive experiments on both public datasets and Drama-ReID reveal the advances of introducing both human body semantic features and its uncertainty.
The main contributions of this paper are summarized as follows.

\begin{itemize}
\item A new person re-ID method for the partial and occluded re-ID problem is proposed, which investigates in depth the human semantic region information and achieves state-of-art in public dataset and Drama-ReID.
\item A human segmentation task is associated to the re-ID model to design an entropy-based probability mask scheme, which split the feature map into confident and unconfident regions, providing more flexibility in semantic alignment.
\item We introduce a semantic alignment using both semantic segmentation information and uncertainty information. The alignment automatically compares the common visible and confident semantic features.
\item We construct a challenging large-scale re-ID dataset of content videos, which is by far the largest dataset focusing on partial and occluded person re-ID.
\end{itemize}

\section{Related Works}\label{sec:related_work}
Deep learning methods \cite{kalayeh2018SPReID,liu2018pose,quan2019Auto-ReID,quispe2019SSP,qi2018maskreid,sun2018PCB} currently dominate the re-ID research community with significant performance on accuracy.
Recent works \cite{sun2018PCB,zhao2017part-aligned,quan2019Auto-ReID,fu2019HPM} further advance the state of the art on holistic person re-ID problem, through learning part-level features.
For example, Sun et al.\cite{sun2018PCB} uniformly partitions the feature map to several local regions and learn a part-level representation respectively.
Zhao et al.\cite{zhao2017part-aligned} extracts part-level features by attention-based methods rather than grid cells or horizontal strips.
Quan et al.\cite{quan2019Auto-ReID} utilizes a part-aware module to enhance the representation of body structural information.
Fu et al.\cite{fu2019HPM} perform horizontal pyramid partition on feature map to extract enhanced discriminative information of all the scale-specific person parts.
All those approaches\cite{sun2018PCB,zhao2017part-aligned,quan2019Auto-ReID,fu2019HPM} have made some progress on holistic datasets\cite{datamarket,dataduke1,dataduke2}.
But an obvious problem with those methods is that their partitions of body are mainly spatial grid, rather than fine-grained and in terms of human semantic.
Therefore, some methods\cite{song2018mask-guided,qi2018maskreid,kalayeh2018SPReID,quispe2019SSP} based on semantic parsing are proposed.
Song et al.\cite{song2018mask-guided} and Qi et al.\cite{qi2018maskreid} employ human mask information to facilitate person re-id models, where the mask-guided map can help 
to remove the background clutters.
Kalayeh et al.\cite{kalayeh2018SPReID} utilizes an extra semantic segmentation network to harness local visual cues to learn the feature for each semantic piece.
Then, they assemble the final discriminative representations with those semantic-level features.
Quispe et al.\cite{quispe2019SSP} combines the semantic parsing and saliency cues to improve the performance of person re-ID model.
Although those methods employ the human semantic information, the representations of different semantic parts are not explicitly aligned and compared, which make them powerless to face the problems of missing parts and occlusion.

In order to solve the partial and occluded re-ID problems, some researchers have proposed several methods based on part mathing\cite{zheng2015AMC,he2018DSR,he2018SFR,sun2019VPM,miao2019PGFA}.
Zheng et al.\cite{zheng2015AMC} proposes a local patch-level matching model called Ambiguity-sensitive Matching Classifier (AMC) and introduces a global part-based matching model called Sliding Window Matching (SWM).
He et al.\cite{he2018DSR,he2018SFR} proposed a series of alignment-free methods that employ sparse feature reconstruction learning, called Deep Spatial Feature Reconstruction(DSR)\cite{he2018DSR} and Spatial Feature Reconstruction(SFR)\cite{he2018SFR} respectively.
Sun et al.\cite{sun2019VPM} proposed a visibility-aware part model (VPM) which learns to locate the visible regions through self-supervision on pedestrian images.
VPM conducted a region-to-region comparison within their shared regions to suppresses the spatial misalignment.
Miao et al.\cite{miao2019PGFA} utilized pose landmarks as a guidance to construct the global feature and then combined the partial features for representation matching.
Based on those related work, we could find that one of the main trends to solve the partial or occluded re-ID is to incorporate the additional information such as body pose or local parts.
But one of the potential problems is that the local representations in those methods lack a correspondence with human fine-grained semantic parts, which is not aligned semantically.
The other problem is the uncertainty of the addition information involved, such as the errors of the pose landmarks or the semantic segmentation.
This problem will be more important when we incorporate the information in the multi-task way, which is more competitive in computation load.
Our method will try to solve the issues by combining both the human semantic-level alignment and the semantic segmentation uncertainty.

\section{Proposed Approach}

\begin{figure*}
\centering
\includegraphics[width=\textwidth]{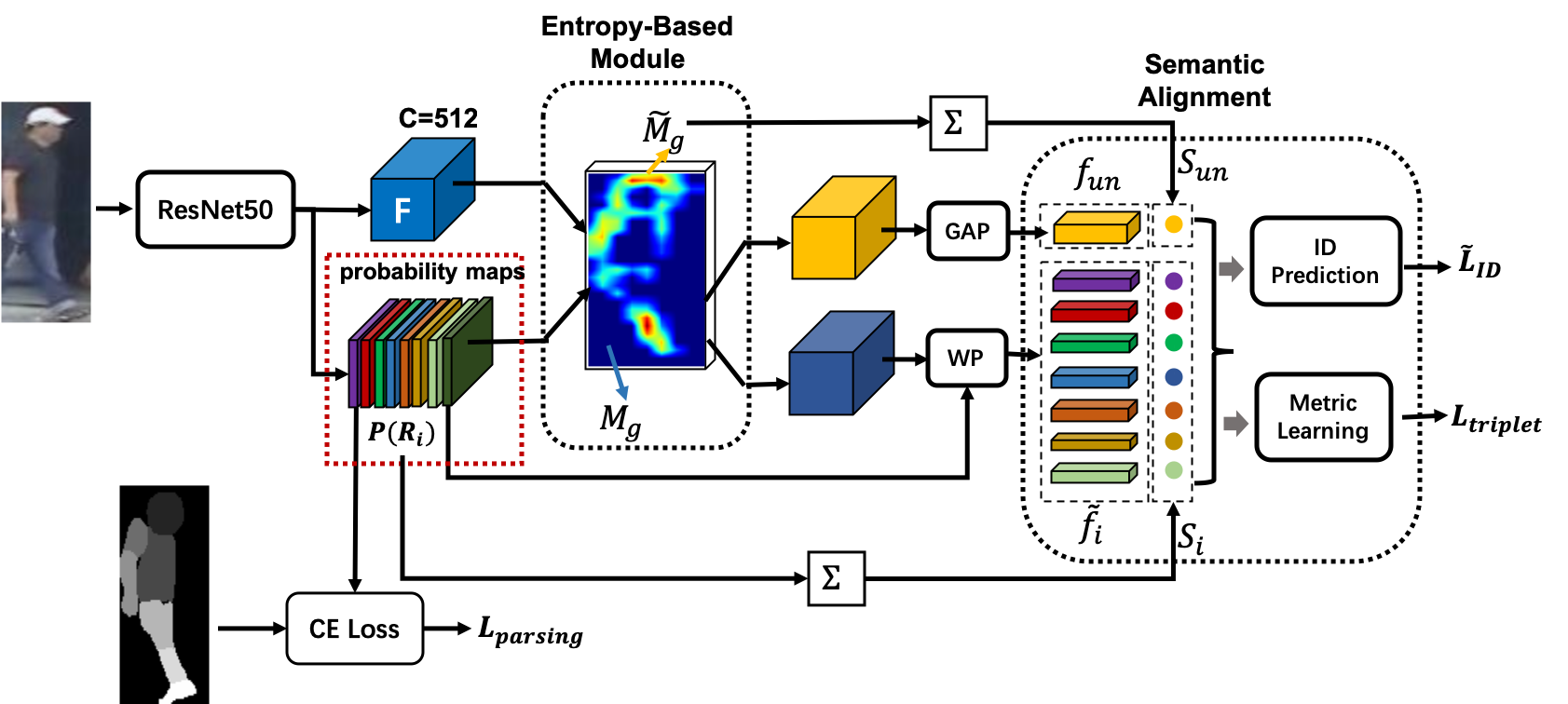}
\caption{Overall structure of the proposed network, where "GAP" means global average pooling and "WP" represents the sum operation on the results of element-wise product between the probability and features defined in Eq.\ref{eq:local_feature}}
\label{fig:overall_structure}
\end{figure*}

Our proposed model consists of three components as shown in Fig.\ref{fig:overall_structure}.
Similar to the traditional re-ID network, the input image will first be sent to the backbone network to obtain the basic features. 
Then a simple human parsing task is carried out based on the feature map of the backbone.
This simple task will, on the one hand, regularize the backbone network to differentiate the different human body parts and the background, on the other hand, provide semantic segmentation for the following modules.
The entropy-based masking module takes as input the semantic segmentation probability of human parsing task and the backbone feature map.
Using the entropy of each semantic segmentation region, an entropy-based mask is generated, separating the feature map into confident features and unconfident features. 
We use the confident features in semantic part to do alignment while incorporating the unconfident features in the training part in an adversarial way to further boost the performance.
Finally, the semantic alignment module aligns the semantic parts of the two people with both high visible probability and high confidence.

\subsection{Backbone Network with Segmentation Task}
We use ResNet50\cite{he2016ResNet} as our backbone network to extract basic feature maps of the given input image.
We remove the average pooling layer and fully connected layer.
Motivated by\cite{sun2018PCB}, the stride of conv4\_1 is set 1 to get larger feature map. 
Formally, we denote the feature map extracted from backbone as $T \in R^{h\times w \times c}$, in which $h$,$w$ and $c$ denote the height, width, channel number respectively.
We append a simple human parsing subtask following the feature map.
This subtask takes as input the backbone output feature map $T$ and then follows one $1 \times 1$ convolution layer and a SoftMax layer.
It outputs $p(R_i|g)$, which denotes the probability of $g$ belong to the human semantic body part $R_i, i = 1,2,...N$.
$N$ is the number of human semantic parts similar to the human parsing domain \cite{gong2017LIP}.
Without loss of generality, $N-1$ means the background. 
$g$ denotes the pixel vector of $T$ (there are $h \times w$ number of $g$ s in $T$), which is a $c$-dim vector.
Parallel to the human semantic parsing task, we add another $1 \times 1$ convolution after the feature map $T$ to reduce the dimension in channels.
This feature map, which is denoted by $F\in R^{h \times w \times c_{new}}$, will be used as the following modules' input.

\subsection{Entropy-based Masking Module}
As discussed in section \ref{sec:related_work}, we will utilize the human body semantic information to solve the occlusion or missing body part issues, which is also the main idea of several related works\cite{kalayeh2018SPReID,quispe2019SSP}.
However, none of them considered the effects of inaccuracy in human semantic segmentation.
When the model depends on the result of segmentation task, the small inaccuracy at the feature map of higher layer might be enlarged in the raw image.
Here we introduce an entropy-based mask to get both confident and unconfident feature part. 
Both of them will provide more flexibility for the following network to utilize the features to align the features semantically.


\begin{figure*} [htbp]
    \centering
    \label{fig:structure_detail}
    \subfigure[]{
    \begin{minipage}[h]{0.45\linewidth}
    \label{subfig:entropy_module}
    \centering 
    \includegraphics[width=3.0in]{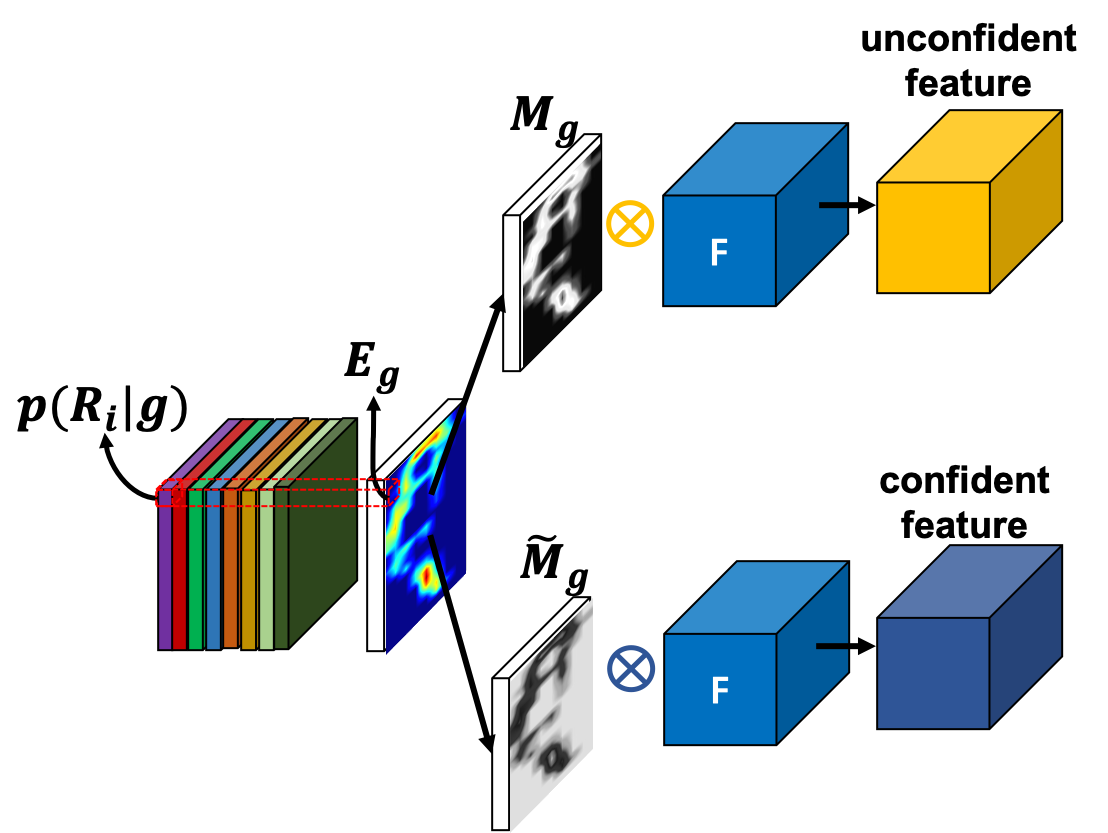} 
    \end{minipage}
    }
    \hspace{10mm}
    \subfigure[]{
    \begin{minipage}[h]{0.45\linewidth} 
    \label{subfig:semantic_align}
    \centering 
    \includegraphics[width=3.0in]{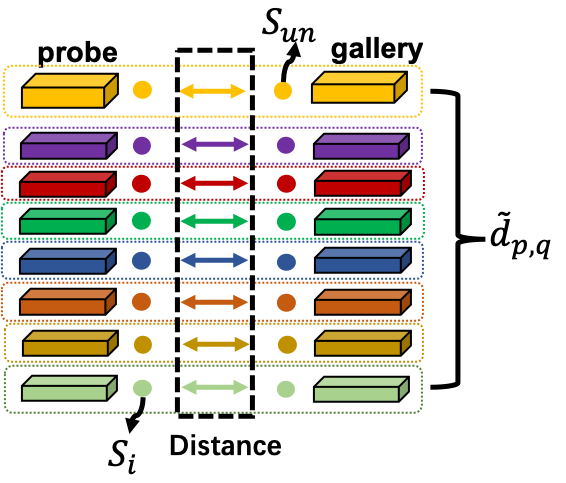} 
    \end{minipage}
    }
    \caption{The detail of the entropy-based module and semantic alignment.}
\end{figure*}

The human segmentation task would output the probability $P(R_i|g)$ of each pixel in the feature map $T$ belonging to semantic part $R_i$.
However, we could not guarantee the accuracy of this task, especially during the training process.
To evaluate the uncertainty of the segmentation result, we use the entropy of segmentation results for $g$, denoted by $E_g$ as follows:
\begin{align}\label{eq5}
E_g=-\sum_{i=0}^{N}{p(R_i|g)\log p(R_i |g)}
\end{align}
If $E_g$ is small, it means that the model can identify the corresponding part of human body with low uncertainty. In other words, $g$ would be the true semantic features 
predicated by $P(R_i|g)$ with high confidence.
$E_g$ is bounded by $E_{max} = \frac{1}{N} \log N$.
Then the entropy mask $M_g$ for unconfident feature can be generated by choosing fixed threshold $\tau$ as follows:
\begin{align}\label{eq:mask_1}
M_g=\begin{cases}
E_g/E_{max}, &E_g/E_{max} >= \tau \\
0, &\text{otherwise}
\end{cases}
\end{align}
Oppositely, we could also get the confident semantic attention map $\tilde M_g$ as 
\begin{align}\label{eq:mask_2}
\tilde M_g=1-E_g/E_{max}
\end{align}

Performing an element-wise multiplication on the feature map $F$ with both entropy mask $M$ and entropy attention map $\tilde M$, we could get the unconfident entropy features and high distinctive semantic features respectively, which is shown in Fig.\ref{subfig:entropy_module}.
This entropy-based mask plays an important role in distinguishing reliable semantic features as well as in the following alignment module.
In general, the unconfident part generally involves the human body intersection part and human profile, since the human parsing model often confuses around the human's nearby body parts or human boundary to background.
Thus, by masking out the unconfident feature map, the human sematic features will be more accurate and helpful for the semantic alignment for people with different visible body part in re-ID.
The entropy-based mask scheme also helps in the model training process, which will be discussed in next session.
Performance comparison and visualization can be found in Section \ref{sec:ablation} and \ref{sec:visualization}.

\subsection{Semantic Alignment}
This section describes the semantic alignment part, which will take each semantic part feature (exclude the background) to form the final feature for re-ID.
Note that for the occlusion case in re-ID, the figure pair is not aligned, i.e. the persons in the image pair have different body parts visible.
Thus, the alignment should use the features corresponding to the shared body parts.
As mentioned in Section \ref{sec:related_work}, some work uses fixed partition of feature map, which is coarse. Other works, although utilize the semantic information, neglect the uncertainty of semantic segmentation which might affect the final performance.


Now we have semantic body part features based on the feature map $F$ and semantic segmentation probability $P(R_i|g)$. Then denote the local features of each semantic region as $f_i$, which is calculated by
\begin{align}\label{eq:local_feature}
f_i=\sum_{g\in R_i} p(R_i|g) g_f,\forall i \in \{1,2,...,N\}
\end{align}
where $g_f$ is the pixel features in $F$ spatially corresponding to the $g$ in feature map $T$.
$F$ and $T$ have same spatial dimensions. 
It is the weighted sum of all the candidate pixels in $F$ using probability denoted by "WP" (weighted pooling) in Fig.\ref{fig:overall_structure}.
Note that Eq.\ref{eq:local_feature} does not consider the segmentation uncertainty, i.e.entropy mentioned in Section \ref{fig:semantic_entropy}. 
We further extend the Eq.\ref{eq:local_feature} as follows
\begin{align}\label{eq:extend_local_feature}
\tilde f_i=\sum_{g\in R_i} \tilde M_g p(R_i|g) g_f,\forall i \in \{1,2,...,N\}
\end{align} 
which means we enhance the confident semantic features and suppress the unconfident semantic features.
Besides, we introduce the visible score $S_i$ similar to \cite{sun2019VPM} for each region defined by $S_i=\sum_{g\in R_i}P(R_i|g)$. 
If a human semantic region is invisible, occluded or non-significant, the visible score will be very small.
Combining both the confident local features and visible score, we could formulate the distance of the persons in image $p$ and $q$ with semantic alignment as
\begin{align}\label{eq:align}
d_{p,q} = \sum_{i=1}^{N-1}S_i^p\cdot S_i^q\cdot D(\tilde f_i^p,\tilde f_i^q)
\end{align}
where $D$ is a distance metric; $N-1$ means we do not select the background region indexed by $N$. 
We can see that if any semantic part is unavailable in either of $p$ or $q$, $S_i^q \cdot S_i^q$ will be close to zero, thus the distance will not consider that part.

We also define the unconfident part feature as $f_{un}$, which can be obtained by multiplying $M_g$ on feature map $F$ with global average pooling denoted by "GAP" in Fig.\ref{fig:overall_structure}. 
We treat $f_{un}$ as the feature for a special semantic region, which contains the uncertain semantic region. 
This type of region, which will be shown in Section \ref{sec:training}, is generally the intersection of body parts and profile. 
Analogue to the semantic region, define the confident score $S_{un}$ for this region as $S_{un} = \sum_g M_g$, we could get an extended generic distance $\tilde d_{p,q}$ combining both the confident features and unconfident feature:
\begin{align}\label{eq:ext_distance}
\tilde d_{p,q}=\frac{\sum_{i=1}^{N-1}S_i^p\cdot S_i^q\cdot D(\tilde f_i^p,\tilde f_i^q)+S_{un}^p\cdot S_{un}^g\cdot D(f_{un}^p,f_{un}^g)}{\sum_{i=1}^{N-1}S_i^p\cdot S_i^q+S_{un}^p\cdot S_{un}^q}
\end{align}

Adding the unconfident region feature into the distance metric will play an adversarial regularization role, which help to stabilize the training and improve the final performance. 
At the beginning of training, we could not get accurate human parsing result. 
The model is more dependent on the unconfident feature, which might include almost all features of the feature map. 
As the training going on, the human parsing task will be more accurate and the semantic part will be enforced by the supervision signal. 
The meaningful semantic region with high certainty will gradually dominate the distance metric.
It's like a self-adversarial game that maintains a dynamic balance between semantic features and high entropy features.
We will investigate it experimentally in Section \ref{sec:ablation}.

\subsection{Network Losses}
Several losses are defined considering the complicate network structure with both entropy-based mask and semantic alignment.
In principle, we use traditional identification loss (ID loss) for each semantic region (including the unconfident region defined above) and triplet loss on the entropy based semantic aligned distance.
For each semantic region $R_i$ except the background, we append the local feature $f_i$ for $R_i$ with one FC layer and one SoftMax layer to predict the people ID $\hat y_{f_i}$.  
The ID loss of semantic regions is defined as
\begin{align}
L_{ID} = \sum_{i}^{N-1} S_i\cdot CE(y, \hat y_{f_i})
\end{align}
where $CE$ is the cross-entropy loss for people id classification, $S_i$ is the visible score for semantic region $i$.
Similarly, we also append one FC layer and one SoftMax layer after the unconfident features $f_{un}$. 
The extended loss considering the unconfident semantic region is 
\begin{align}
\tilde L_{ID} = \sum_{i}^{N-1} S_i\cdot CE(y, \hat y_{f_i}) + S_{un} \cdot CE(y, \hat y_{un})
\end{align}
where $\hat y_{un}$ is the ID prediction based on the unconfident feature. 
Note that confident scores for each semantic region even including the unconfident semantic region are involved in those loss definitions.
Also, for the confident regions, we use $f_i$ defined in Eq.\ref{eq:local_feature} rather than $\tilde f_i$ defined in Eq.\ref{eq:extend_local_feature}, expecting the loss has strong supervision on the semantic segmentation.

For the triplet loss\cite{hermans2017triplet}, it is defined as 
\begin{align}
L_{triplet} = \max \{0, \tilde d_{a, p} - \tilde d_{a, n} + m \}
\end{align}
where $a$ and $p$ are the positive pair while $a$ and $n$ represent negative pair; $\tilde d$ is the extended distance defined in Eq.\ref{eq:ext_distance}. $m > 0$ is the margine the triplet loss wants to guarantee between positive and negative pairs. 
The final loss is written as 
\begin{align}\label{eq:total_loss}
L_{all} = \lambda L_{parsing} + \tilde L_{ID} + L_{triplet}
\end{align}
where $L_{parsing}$ is the general segmentation loss for the human segmentation task, which is not expanded in detail for brevity. 
$\lambda$ is the weight for the human segmentation task.
Because human parsing task is important for our model, we will investigate the chosen of $\lambda$ in ablation study.

\section{Experiments}\label{sec:experiment}
\subsection{Datasets and Evaluation Measures}
To demonstrate the performance of our method on re-ID problem with occlusion, missing part in both traditional surveillance and video content case, we evaluate the proposed model on dataset with both partial re-ID dataset and holistic re-ID datasets. However, there is few large data for re-ID problem with occlusion, missing part, especially in video content domain. We construct our own data set called Drama-ReID.

\begin{figure*}
    \centering
    \includegraphics[width=\textwidth]{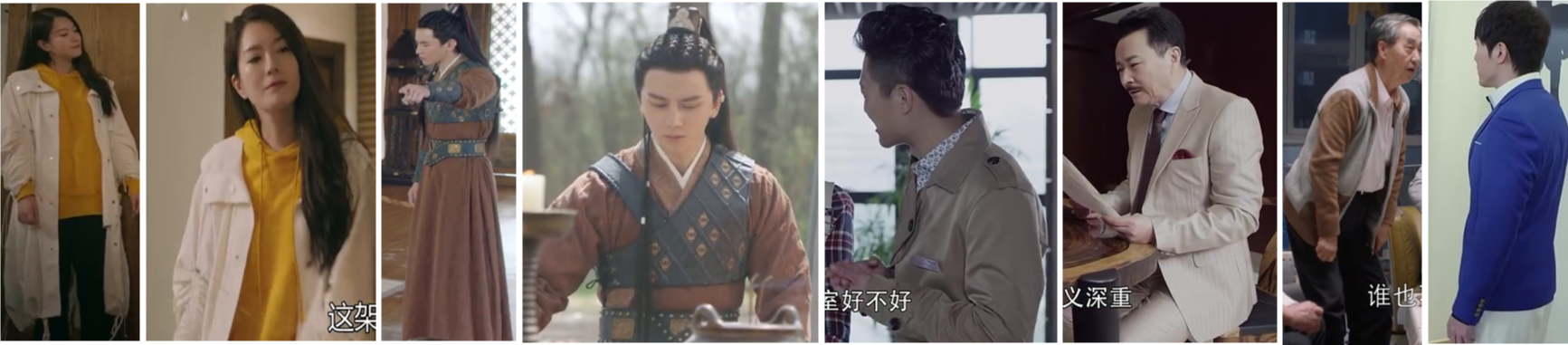}
    \caption{Examples of Drama-ReID Dataset}
    \label{fig:dramaset}
\end{figure*}

\textbf{Drama-ReID}.
Drama-ReID is a dataset in content video domains, i.e. the human figures are extracted from the TV and movies.
As we discussed before, the content video is artifact, which is different from the traditional surveillance case. 
The people in the figure will be captured with different shot languages such as close-up shot, full-shot, different angles, etc.
We extracted $\sim$300,000 images of $\sim$10,000 identities from $\sim$500 different TV shows with existing human Yolo detector \cite{redmon2016yolo}.
Each identity has at least 30 pictures with different poses, shot angle, half-full body’s, etc.
Fig.\ref{fig:dramaset} shows several examples of Drama-ReID.
Actors with different clothing will be treated as different identities.
As far as we know, this dataset is now the largest re-ID dataset in both content video and surveillance video.
We will release it in the near future to inspire research insights on complicated re-ID cases with occlusions, missing parts, angles, poses scale variations, etc.

\textbf{Partial-REID\cite{datapartialreid} and Partial-iLIDS \cite{datapartialilids}}.
Partial-REID and Partial-iLIDS are typical evolution dataset for occlusion and missing part research in re-ID.
Partial-REID includes 600 images from 60 people, with five full-body images and five partial images per person. We follow the evaluation protocols in \cite{zheng2015eval} where 300 full-body images of 60 identities are used as the gallery set and 300 occluded-body images of the same 60 identities are used as the probe set. 
Partial-iLIDS is derived from iLIDS \cite{datapartialilids}, which is collected in an airport and the lower-body of a pedestrian is frequently occluded by the luggage. The iLIDS dataset contains a total of 476 images of 119 people captured by multiple non-overlapping cameras.

\textbf{Market-1501\cite{datamarket} and DukeMTMC-reID\cite{dataduke1,dataduke2}}.
We also evaluate our model in public full-body re-ID dataset. 
Market-1501 contains 1501 identities, which is extracted from 6 camera viewpoints, 19732 gallery images and 12936 training images detected by DPM\cite{felzenszwalb2009DPM}.
DukeMTMC-reID contains 1404 identities, 16522 training images, 2228 queries, and 17661 gallery images.

\textbf{Evaluation Metrics}.
We use Cumulative Matching Characteristic (CMC) curves and mean average precision (mAP) to evaluate the quality of different person re-ID models. All the experiments are performed in a single query setting. 
However, for Drama-ReID, we also require the evaluation metric PR-AUC\cite{AUC}, which is defined as the area under the P/R curve.
Because the re-ID problem is in essence a binary classification problem, PR-AUC will give us more insights about the model performance than ranking metric. What's more, in real application, we commonly use re-ID to matching tracking results in different shot in content video. Thus, AUC provides a guidance to select the threshold to tradeoff between precision and recall.

\subsection{Implementation Details}\label{sec:training}
Our backbone is ResNet50\cite{he2016ResNet} without the average pooling layer and fully connected layer.
At the same time, the stride of conv4\_1 is set to 1 in order to obtain lager feature map.
We initialize the backbone model by the ImageNet\cite{deng2009imagenet} pre-trained model. 
In our experiment setting, the input image is resized to $384 \times 128$
and augmented by random flipping and random erasing\cite{zhong2017REA}. 
We use batch hard triplet loss\cite{hermans2017triplet} and set batch size to 64 with 8 identities (8 images per identity). 
We use standard Stochastic Gradient Descent (SGD) optimizer, with the initial learning rate being 0.1 and decaying to 0.01 after 30 epochs. 
The proposed method is trained with 70 epochs.
Before training, we process the images using human parsing model \cite{gong2017LIP}, which gives 20 semantic labels.
By combing the minor human semantic part, we get 8 typical parts, namely background, head, torso, upper-arm, lower-arm, upper-leg, lower-leg and foot.
At the same time, we also train a strong baseline model which only has the Resnet50 followed by the hard triplet loss.

\subsection{Result Comparisons}
\subsubsection{Results on Drama-ReID.} 
Table.\ref{table:drama_reid_result} shows the result of our model compared with several existing works. 
Our model achieves 96.7\% Rank-1 accuracy and 91.3\% mAP, which outperforms all the previous methods.
At the same time, to compare the PR-AUC performance with existing work, we also trained some models following the authors' implementations and official opensource code on Drama-ReID. 
The PR-AUC of our model is 0.86, which is much better than them.
\begin{table}
\begin{center}
\caption{Results on Drama-ReID dataset}
\label{table:drama_reid_result}
\setlength{\tabcolsep}{1mm}{
\begin{tabular}{c|c|c|c}
\hline
 Method& Rank1 & mAP & PR-AUC \\
 \hline
 \hline
 DSR(CVPR2018)\cite{he2018DSR}& 82.5 &73.0 &0.47 \\
 SFR\cite{he2018SFR}& 86.9 &76.3&0.60 \\
 PCB(ECCV2018)\cite{sun2018PCB}& 93.8 &84.5 &0.79 \\
 VPM(CVPR2019)\cite{sun2019VPM}& 94.5 &86.7&0.81 \\ \hline
 Baseline& 93.2 &83.6 & 0.78 \\ 
 ESA-ReID(ours)& \textbf{\textcolor{red}{96.7}} & \textbf{\textcolor{red}{91.3}} & \textbf{\textcolor{red}{0.86}} \\
\hline
\end{tabular}
}
\end{center}
\end{table}

\subsubsection{Results on Partial-REID\cite{datapartialreid} and Partial-iLIDS\cite{datapartialilids}.}
To compare the performance on partial re-ID case, 
we compare our model with several existing partial person re-ID methods, 
including MTRC\cite{liao2012MTRC}, AMC+SWM\cite{zheng2015AMC},
DSR\cite{he2018DSR}, SFR\cite{he2018SFR}, 
STNReID\cite{luo2020stnreid},VPM\cite{sun2019VPM} and PGFA\cite{miao2019PGFA}.
Same to the previous works, we train our model on Market-1501 training set. 
As shown in Table.\ref{table:partial_reid_result}, the performance of our model is quite competitive. 
The Rank-1/Rank-3 of our method achieves 78.6\%/84.3\% and 73.6\%/82.4\% on Partial-REID and Partial-iLIDS.
Comparing to the strongest competing method PGFA\cite{miao2019PGFA}, our method surpasses it by 10.6\% on Partial-REID and 4.5\% on Partial-iLIDS in Rank-1 respectively, which are a large margin.
The results demonstrate that the entropy-based semantic feature alignment is very useful for partial and occlusion re-ID.
\begin{table}
\centering
\caption{Results on partial re-ID datasets}
\label{table:partial_reid_result}
\setlength{\tabcolsep}{1mm}
{
\begin{tabular}{c|c|c|c|c}
\hline
\multirow{2}{*}{Method}&\multicolumn{2}{c|}{Partial-REID}&\multicolumn{2}{c}{Partial-ILIDS} \\ 
 \cline{2-5}
 &rank1 & rank3 & rank1 & rank3 \\ 
\hline
\hline
MTRC\cite{liao2012MTRC} & 23.7 & 27.3 & 17.7 & 26.1 \\
AMC(ICCV2015)\cite{zheng2015AMC} & 37.3 & 46.0 & 21.0 & 32.8 \\
DSR(CVPR2018)\cite{he2018DSR} & 50.7 & 70.0 & 58.8 & 67.2 \\
SFR\cite{he2018SFR} & 56.9 & 78.5 & 63.9 & 74.8 \\ \hline
STNReID\cite{luo2020stnreid} & 66.7 & 80.3 & 54.6 & 71.3 \\
VPM(CVPR2019)\cite{sun2019VPM} & 67.7 & 81.9 & 65.5 & 74.5 \\
PGFA(ICCV2019)\cite{miao2019PGFA} & 68.0 & 80.0 & 69.1 & 80.9 \\ \hline
Baseline & 55.0 & 76.1 & 56.2 & 66.3 \\
ESA-ReID(ours) & \textbf{\textcolor{red}{78.6}} & \textbf{\textcolor{red}{84.3}} & \textbf{\textcolor{red}{73.6}} & \textbf{\textcolor{red}{82.4}} \\
\hline
\end{tabular}
}
\end{table}

\subsubsection{Results on Market-1501\cite{datamarket} and DukeMTMC-reID\cite{dataduke1}.}
We also test our model on Market-1501\cite{datamarket} and DukeMTMC-reID\cite{dataduke1}.
As shown in Table.\ref{table:market_reid_result}, even on holistic person re-ID datasets, our model achieves comparable performance with state-of-the-art.

\begin{table}
\centering
\caption{Results on Market-1501 and DukeMTMC-reID}
\label{table:market_reid_result}
\setlength{\tabcolsep}{1mm}{
\begin{tabular}{c|c|c|c|c}
\hline
\multirow{2}{*}{Method}& \multicolumn{2}{c|}{Market-1501} &\multicolumn{2}{c}{DukeMTMC} \\ 
\cline{2-5}
&rank1 & mAP & rank1 & mAP \\ 
\hline
\hline
PCB(ECCV2018)\cite{sun2018PCB} & 93.8 & 81.6 & 83.3 & 69.2 \\
SPREID(CVPR2018)(\cite{kalayeh2018SPReID} & 92.5 & 81.3 & 84.4 & 71.0 \\
HPM(AAAI2019)\cite{fu2019HPM} & 94.2 & 82.2 & 86.6 & 74.3 \\
Auto-REID(ICCV2019)\cite{quan2019Auto-ReID} & 94.5 & 85.1 & \textbf{\textcolor{red}{88.5}} & 75.1 \\ \hline
DSR(CVPR2018)\cite{he2018DSR} & 83.5 & 64.2 & - & - \\
SFR\cite{he2018SFR} & 93.0 & 81.0 & 84.8 & 71.2 \\
VPM(CVPR2019)\cite{sun2019VPM} & 93.0 & 80.8 & 83.6 & 73.6 \\
PGFA(ICCV2019)\cite{miao2019PGFA} & 92.4 & 77.3 & 82.6 & 65.5 \\
STNReID\cite{luo2020stnreid} & 93.8 & 84.9 & - & - \\ \hline
Baseline & 93.2 & 82.5 & 84.9 & 75.4 \\
ESA-ReID(ours) & \textbf{\textcolor{red}{94.5}} & \textbf{\textcolor{red}{86.3}} & {87.7} & \textbf{\textcolor{red}{77.6}} \\ \hline
\end{tabular}
}
\end{table}

\subsection{Ablation Study}\label{sec:ablation}
In this section, we do ablation study on several components in our proposed model. 

\subsubsection{The Impact of Coefficient $\lambda$.}
\begin{figure} 
    \centering 
    \includegraphics[width=3.0in]{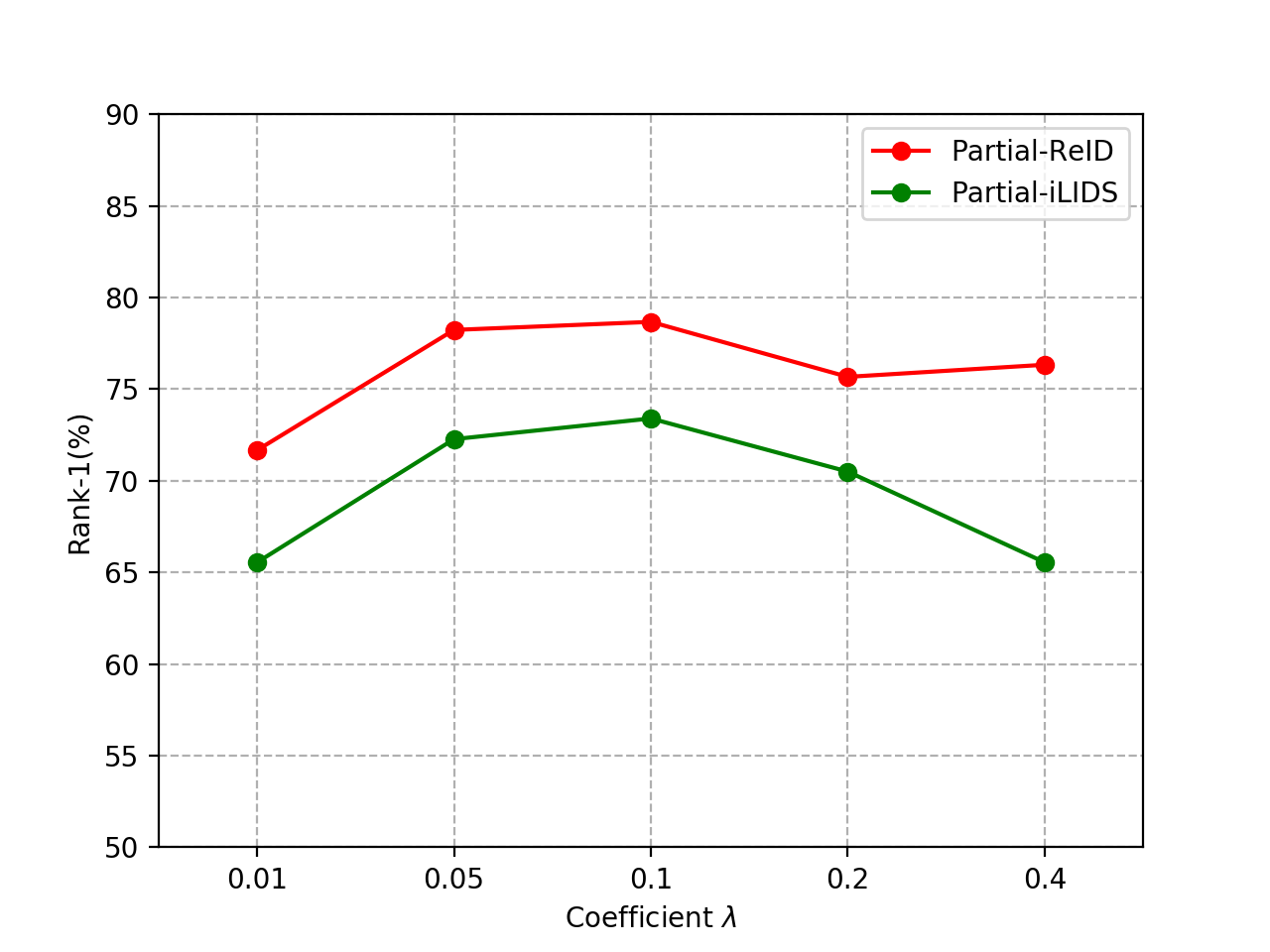} 
    \caption{Results with different $\lambda$ s} 
    \label{fig:lambda} 
\end{figure}

The loss function defined in Eq.\ref{eq:total_loss} has a hyper parameter $\lambda$. 
It's the weight on human semantic parsing task, which is important in our task.  
We conduct a series of experiments with different $\lambda$ s to select the best choice.
As shown in Fig.\ref{fig:lambda}, when $\lambda$ is small (for example 0.01), the performance of semantic parsing task is difficult to improve, because incorrect semantic matching will reduce the final performance of the model. 
When $\lambda$ is too large, the optimizer will favor too much on semantic parsing task rather than semantic alignment and matching task, which will also lead a poor performance on the representation learning.
For the datasets Partial-REID and Partial-iLIDS, we best $\lambda$ is our experiment is 0.1.

\subsubsection{The Impact of Threshold $\tau$.} 
\begin{figure} 
    \centering 
    \includegraphics[width=3.0in]{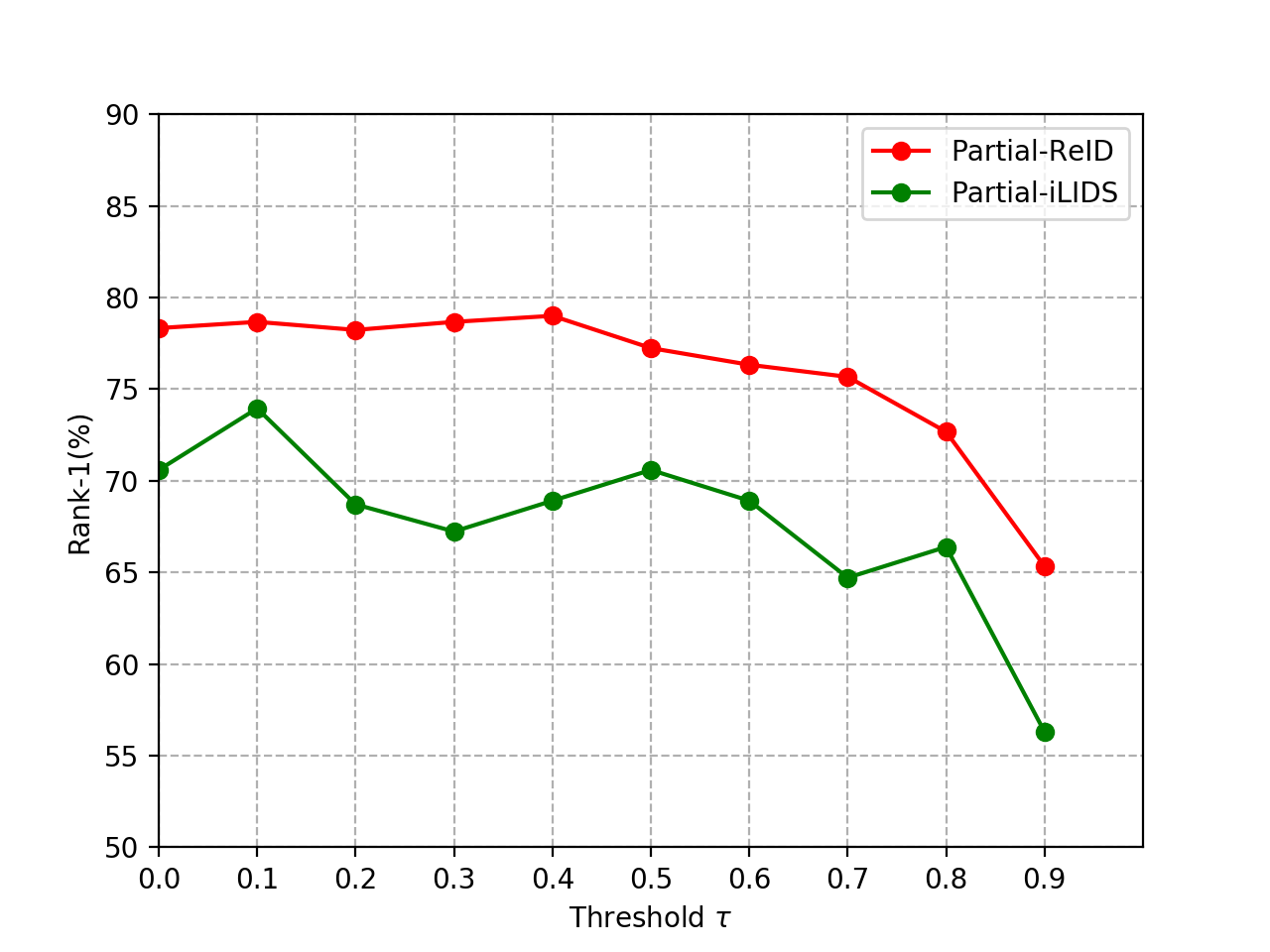} 
    \caption{Results with different $\tau$ s} 
    \label{fig:tau} 
\end{figure}

In Eq.\ref{eq:mask_1} and Eq.\ref{eq:mask_2}, $\tau$ is the threshold of the entropy mask. 
It works like a gate. During each train step, it will select the confident parts in the semantic segmentation region.
As shown in Fig.\ref{fig:tau}, we increase $\tau$ from 0.1 to 0.9. 
When $\tau$ is small, such as 0.1, the gate will keep large number of unconfident features.
It has high chance to lose useful human semantic features and the unconfident feature will be overwhelmed, which won’t help for the final result.
On the other side, when $\tau$ is large, many regions with high uncertainty will be involved in the semantic regions, degrading the alignment performance based on the semantic regions.


\subsubsection{The Impact of Unconfident Features.}
We further investigate the performances if we remove or change the unconfident features.
First, we replace the unconfident feature by global feature, i.e. removing the entropy-based mask and the unconfident features in the compare distance in Eq.\ref{eq:ext_distance}.
The model actually degenerates to a model that does alignment with semantic segmentation feature and the global feature. 
We denote the result as "g" in Table \ref{table:insalience_analysis}.
Second, we remove the unconfident feature part and only use semantic segmentation features, whose result is denoted by "w".
Third, we design a dynamic threshold experiment in entropy-based mask generation.
We only choose features larger than the largest half part of entropy as the unconfident region, whose result is denoted by "d".
By comparing the results of "g" and "d" with our model, we could find the advantage of the unconfident region features by the entropy mask.
The results demonstrate the adversarial regularization effects of incorporating unconfident features into the alignment distance as mentioned in Section \ref{fig:overall_structure}.
The incorporation of unconfident features in semantic alignment will stabilize the training process and boost the final performance in a self-adversarial way.
The results of "d" and ESA-ReID indicate that a fixed $\tau$ is preferable under this self-adversarial regularization settings.

\begin{table}
\begin{center}
\caption{Results on unconfident region features}
\label{table:insalience_analysis}
\setlength{\tabcolsep}{1mm}{
\begin{tabular}{c|c|c|c}
\hline
\multirow{2}{*}{Setups} & Partial-REID & Partial-iLIDS & Drama-ReID \\
 \cline{2-4}
 & rank1 & rank1 & PR-AUC \\ 
 \hline
 \hline
 g & 76.33 &70.59 & 0.84  \\
 w & 75.31 &65.79 & 0.79 \\
 d &76.00	&66.39 & 0.81 \\
 our & \textbf{\textcolor{red}{78.6}} & \textbf{\textcolor{red}{73.6}} & \textbf{\textcolor{red}{0.86}} \\
\hline
\end{tabular}
}
\end{center}
\end{table}

\begin{figure*} 
    \centering 
    \includegraphics[width=4.5 in]{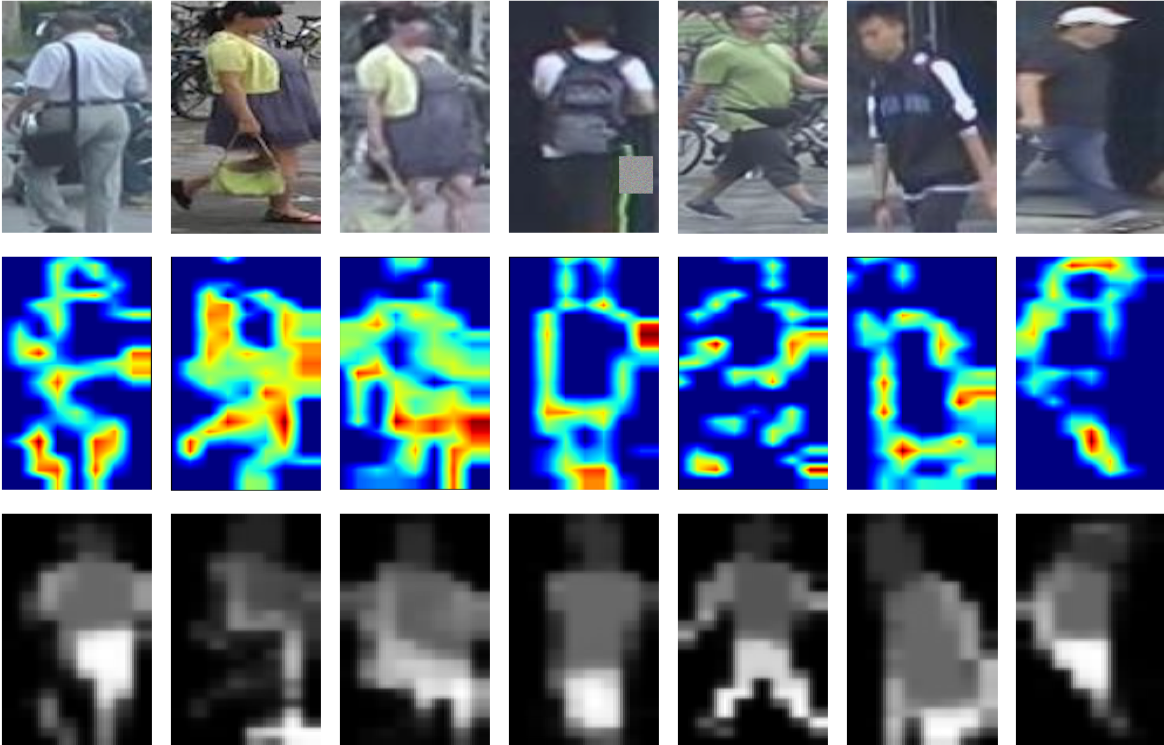} 
    \caption{Visualization of the semantic parsing result and entropy score of person images. 
    The 1st row shows input person images; the 2nd row is the corresponding entropy score map; the 3rd row is the corresponding semantic parsing results.} 
    \label{fig:semantic_entropy} 
\end{figure*}

\subsection{Visualization}\label{sec:visualization}
To see more clearly the entropy value of different regions, we visualize the semantic parsing result and entropy score of the person images.
As shown in Fig.\ref{fig:semantic_entropy}, we found that the high entropy region appears mainly at the edges of human semantic parts and intersection of semantic regions. 
Those regions are difficult to determine exactly which semantic part they belong to. 
If we force those regions simply by probability to certain semantic part regions, those incorrectly regions will contaminate the semantic features.
On the other side, we can see that the unconfident part has some other semantic part which is not common, such as the 5th person’s shoes and the 7th person’s hat. 
Those minor but distinguishable region is not involved in the general semantic regions. 
But by incorporating the entropy-based mask and unconfident region feature, 
those features can also be utilized in our model.

\section{Conclusions}
In this paper we propose an entropy-based semantic alignment model.
This model utilizes the human semantic region information by a simple human parsing task.
The inaccuracy of human semantic parsing is innovatively utilized by an entropy-based mask scheme. 
The semantic alignment will consider both the visible score and confident score of each semantic part, which is expected to handle the case of occlusion and partial re-ID.
We also construct the largest re-ID dataset especially for the content videos, which has many cases of partial, occlusion, shot angle variations. 
Our model shows prominent results on both our new dataset and existing public datasets.
However, there is still many other unsolved cases such as angle variations, large pose-variations in re-ID, which can be seen in our new dataset. 
We will release the dataset, hoping to inspire more insights on the re-ID in the domain of content video.


\clearpage
%
%
\bibliographystyle{ieee_fullname}
\bibliography{egbib}
\end{document}